\documentclass[journal]{IEEEtran}
\usepackage{amsmath,graphicx,caption,subcaption,bbm,array,float,hyperref}
\newcolumntype{P}[1]{>{\centering\arraybackslash}p{#1}}

\begin{document}

\title{Joint Neural Networks for One-shot Object Recognition and Detection}

\author{Camilo J. Vargas,
        Qianni Zhang,
        Ebroul Izquierdo.
\thanks{C. J. Vargas, Q. Zhang and E. Izquierdo are with the School of Electronic Engineering and Computer Science, Queen Mary University of London, London. UK.
E-mail: \{c.j.vargascortes, qianni.zhang, ebroul.izquierdo\}@qmul.ac.uk\protect\\
}}

\maketitle

\begin{abstract}
This paper presents a novel joint neural networks approach to address the challenging one-shot object recognition and detection tasks. Inspired by Siamese neural networks and state-of-art multi-box detection approaches, the joint neural networks are able to perform object recognition and detection for categories that remain unseen during the training process. Following the one-shot object recognition/detection constraints, the training and testing datasets do not contain overlapped classes, in other words, all the test classes remain unseen during training. The joint networks architecture is able to effectively compare pairs of images via stacked convolutional layers of the query and target inputs, recognising patterns of the same input query category without relying on previous training around this category. The proposed approach achieves 61.41\% accuracy for one-shot object recognition on the MiniImageNet dataset and 47.1\% mAP for one-shot object detection when trained on the COCO dataset and tested using the Pascal VOC dataset. Code available at \url{https://github.com/cjvargasc/JNN_recog} and \url{https://github.com/cjvargasc/JNN_detection/}.
\end{abstract}


\IEEEpeerreviewmaketitle

\section{Introduction}

\IEEEPARstart{T}{he} automatic recognition, description, classification, and grouping of patterns are relevant tasks in a variety of engineering and scientific disciplines such as biology, psychology, medicine, marketing, security, transport, and remote sensing. Regardless of almost 50 years of research, the design of a general-purpose machine pattern recognizer remains an elusive goal~\cite{paper1}\cite{Jref1}.

Deep Neural Networks (DNNs) have achieved the state-of-the-art performance on a variety of pattern recognition tasks, most notably visual classification problems. Typically, a very large amount of training samples are required to achieve state-of-art results~\cite{paper2}. In contrast, real world applications often require to achieve high recognition performance with very limited datasets. In those cases the classic deep neural networks fall short, leading to overfitting on the training sets and poor generalizations on the test sets~\cite{paper3}. Thus, the capability to generalize visual concepts from small amounts of data becomes a highly desired characteristic in modern computer vision systems

This paper introduces Joint Neural Networks to tackle the one-shot object recognition and detection tasks. Here, the Joint Neural Network model is trained as a matching network to perform recognition or detection in pairs of images. The proposed approach is tested using input query and target images belonging to categories that remain unseen during the training stage. Such approach is applicable to any field where data availability may be reduced to a single sample, where traditional machine learning and data augmentation approaches fall short.

Two different architectures akin to Siamese neural networks (SNNs) are presented in this paper to tackle the recognition and detection tasks. Both Joint Neural Network architectures rely on the \textit{joint layer} key concept. The joint layers proposed in this paper are convolutional layers that merge the convolutional filters resulting from the two branches of the Joint Network architectures via concatenation. The outputs of the joint layers are built to match the defined backbone architecture used for the Joint Neural Network.

For recognition, the proposed novel approach is akin to the well known SNN with an AlexNet backbone. Their performance is compared in the experimental section along with several few-shot learning approaches in the context of one-shot recognition. For detection, the proposed recognition approach is adapted inspired by one-stage multi-box detection approaches similar to that introduced by YOLO~\cite{YOLO1}~\cite{YOLO2}. The main contributions of this paper are:\\
- A scalable framework for object one-shot recognition and detection inspired by Siamese Neural Networks and anchor based one-stage multibox detectors, that additionally do not require support sets or additional training/fine-tuning for unseen classes.\\
- The introduction of two novel Joint Neural Networks architectures for one-shot recognition and detection respectively.

\section{Related Work}

Previous to modern machine learning approaches, the initial approaches to engage in the one-shot object detection and recognition problems were based on image processing techniques. The most popular image processing based approaches relied on the localisation and description of interest points or keypoints, as it is the case of SIFT \cite{SIFT} and SURF \cite{SURF} for object detection. Approaches based on keypoint detection and description are often designed to offer feature scale, rotational, and illumination invariance. When applied in realistic scenarios, one of the main withdraws of such approaches is the limited capability of matching feature keypoints under illumination or structural variations in particular when the background scenery is complex.

As the performance of image processing based approaches reached a plateau new machine learning approaches were introduced to address the one-shot problem. The goal of machine learning algorithms is to discover statistical structures within data, but building modern models such as convolutional neural networks (CNNs) from scratch requires a large amounts of labeled training data~\cite{YOLO2}. Following~\cite{paper5}, the seminal work towards one-shot learning dates back to the early 2000’s with work in~\cite{paper6} where the authors developed a variational Bayesian framework for one-shot image classification using the premise that previously learned classes can be leveraged to help forecast future ones when very few examples are available from a given class~\cite{paper6}~\cite{paper7}. However, these approaches are still based on keypoints, inheriting the same drawbacks and the problem remains unsolved for real world scenarios.

Data synthesis and transfer learning are commonly used techniques to support models carrying out the detection or recognition tasks when the labeled training data is limited. Data synthesis focuses on producing artificial data either manually from scratch or using advanced data manipulation techniques to produce novel and diverse training examples.  When creating new exemplars from existing instances, the available data must be enough to render novel and diverse instances to produce novel and diverse training examples~\cite{thesis37}. In general, the data synthesis approach serves well for cases in which at least a few samples of the target categories are available but its effectiveness is limited for one-shot object recognition/detection.

Transfer learning is one of the most relevant tools in machine learning, proposed to solve the insufficient training data problem. It aims to transfer the knowledge from a source domain to a target domain by relaxing the assumption that the training data and the test data must be independent and identically distributed. This leads to a great positive effect on many domains that are difficult to engage with because of insufficient training data~\cite{thesis36}\cite{Jref2}. Fine-tuning remains the method of choice for transfer learning with neural networks in which a model is pre-trained on a source domain (where data is often abundant), the output layers of the model are adapted to the target domain, and the network is fine-tuned via back-propagation. This approach was pioneered in \cite{thesis31lost} by transferring knowledge from a generative to a discriminative model, and has since been generalized with great success \cite{thesis32lost}\cite{thesis32lost2}\cite{thesis32lost3}. Traditionally, this method also requires at least a few samples of the target domain categories to be available in order to perform the fine-tuning training process, which becomes a limiting factor when dealing with one-shot detection or recognition. 

Recently, a set of approaches for few-shot recognition and suitable for one-shot object recognition have been proposed \cite{46fs1}, \cite{46fs2}, \cite{46fs3}, \cite{46fs4}, \cite{46fs5}, \cite{Jref3}. The goal of meta-learning is to train a model on a variety of learning tasks, such that it can solve new learning tasks using only a small number of training samples. The work presented in~\cite{46fs1} proposes a general and model-agnostic meta-learning algorithm. The algorithm is model-agnostic in the sense that it can be directly applied to any learning problem and model that is trained with a gradient descent procedure. The key idea behind the proposed approach is to train the model’s initial parameters such that the model has maximal performance on a new task after the parameters have been updated through one or more gradient steps computed with a small amount of data from that new task.

A set of architectures called Siamese Neural Networks (SNNs) have been also proposed to tackle to one-shot recognition task. A SNN consists of twin neural networks with tied parameters that takes two inputs and join them using an cost function at the top. This function computes a distance metric between resulting the highest-level feature representation on each side. Once trained, SNNs can then capitalise on powerful discriminative features to generalise the predictive power of the network not just to new data, but to entirely new classes from unknown distributions. Furthermore, the shared weights constraint on the twin CNN's guarantees that two extremely similar images could not possibly be mapped by their respective networks to very different locations in feature space given that each network computes the same function~\cite{paper5}.

Siamese Neural Networks have been applied for a wide set of tasks including character recognition~\cite{paper5}, person re-identification~\cite{paper13}, face verification~\cite{paper14} and tracking~\cite{paperref24}; given their balance between accuracy and speed~\cite{paperref23} for one-shot recognition. Traditionally, Siamese networks were trained with the so called “contrastive loss”~\cite{paperref26}~\cite{paperref27}. The main idea here is to construct a Euclidean space where the outputs of the sister networks are close to each other for matching pairs while simultaneously negative pairs are pushed away~\cite{paperref28}.

Although, most of the work found in the literature is focused on one-shot object recognition, recent research works have started pioneering possible solutions to the challenging one-shot detection problem \cite{coAE}\cite{MFCOS} and extending it to fields such as semantic segmentation \cite{SiamRPN} and visual tracking\cite{SiamFC}\cite{CompNet}. In contrast to the work presented in this paper, the aforementioned approaches are based on two-stage object detection approaches.

This work introduces an approach inspired by SNNs to tackle the one-shot object recognition problem, that additionally uses concepts from one-stage detectors such as YOLO~\cite{YOLO2} to extend its application to the one-shot object detection problem. Following the traditional object detection literature case where enough data is available for training, one-stage detectors offer a trade-off between accuracy and speed favorable for real world applications. The novel architecture proposed is based on the idea of allowing the model to Figure out by itself the convolutional activations that lead to pattern matching and locations definition, in contrast to the common encoding comparison at the top of the Siamese sister networks.

\section{Proposed Approach}

The fundamental idea behind the proposed joint neural networks is to allow the convolutional layers to learn the activations that define matches between pairs of images, instead of having a single comparison layer at the top of the architecture as in SNNs.

\begin{figure}
\centering
\includegraphics[width=265pt]{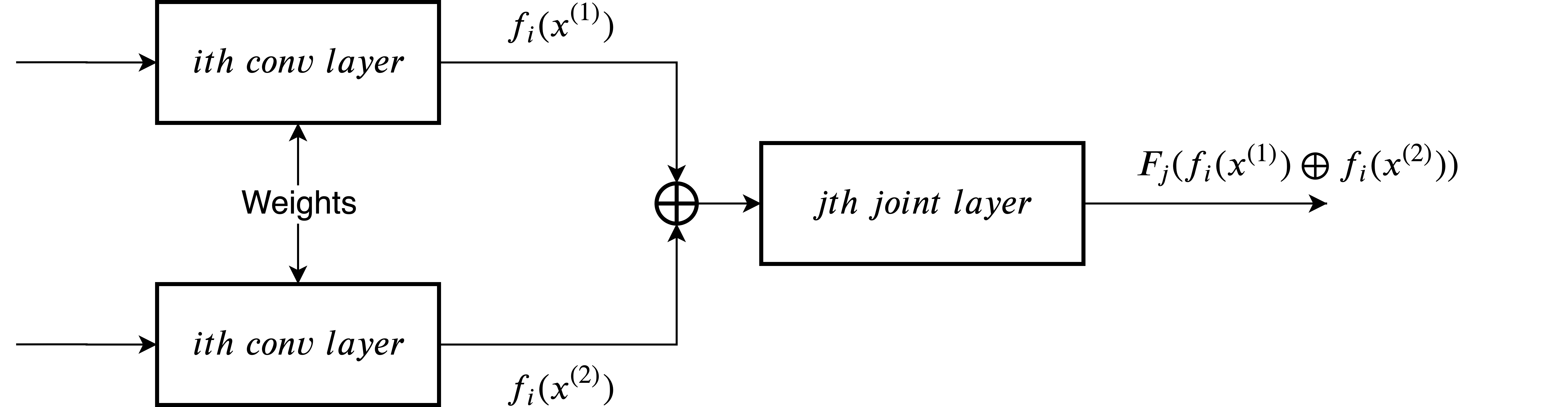}
\caption{Proposed joint layer structure.}
\label{Jlayer}
\end{figure}

For this purpose we define \textit{joint convolutional layers}, which are the backbone of Joint neural networks and can be used within any convolutional neural network architecture. The so called joint layers are convolutional layers like those found in common convolutional neural networks; where, the feature vector taken as input is the result of a concatenation operation between the previous convolutional layers on each branch of the Joint neural network. Figure~\ref{Jlayer} illustrates the structure of a joint layer. The output of the proposed joint layers is defined to match the architecture used as backbone of the Joint neural network. In this work the AlexNet and DarkNet19 architectures are used to tackle the recognition and detection problems respectively.

In Figure\ref{Jlayer} $f_i(x^{(1)})$ and $f_i(x^{(2)})$ are the feature vectors outputted by the $i$th convolutional layer for the first and second branch respectively. $F_j(f_i(x^{(1)}) \oplus f_i(x^{(2)}))$ is the feature vector outputted by the $j$th joint layer. And $\oplus$ is the standard concatenation operation merging the output vectors $f_i(x^{(1)})$ and $f_i(x^{(2)})$.

\subsection{Recognition}

When undertaking the one-shot recognition problem the main goal is to perform pattern matching for a pair of images $x^{(1)}$, $x^{(2)}$; outputting a predicted similarity metric $P$ in the same manner as traditional Siamese neural networks. Given that AlexNet is an architecture commonly used as the backbone of SNNs~\cite{paperref24} we adopt the same base architecture for demonstration and comparison purposes.

\begin{figure*}[ht!]
\centering
\includegraphics[width=0.90\textwidth]{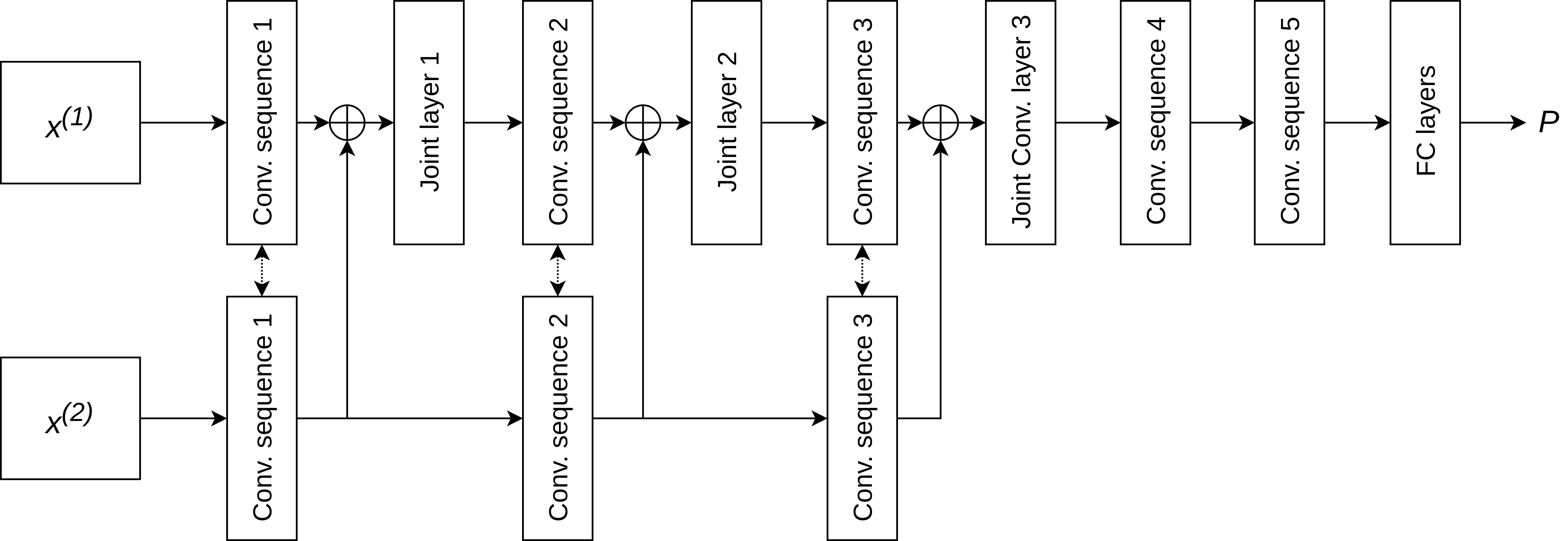}
\caption{Proposed AlexNet-based recognition architecture.}
\label{arch_recog}
\end{figure*}

Figure~\ref{arch_recog} depicts the proposed architecture showing the two branches of the network. The main branch of the architecture at the top of Figure~\ref{arch_recog} is built over the AlexNet architecture with the additional joint layers that will integrate outputs from the second branch at the bottom, thus supporting information sharing between the network branches. The convolutional layers in the secondary branch of the architecture share weights with the main branch, the same way as sister networks in traditional SNN approaches. Additionally, the last fully connected layer of the architecture consists of a single neuron that will define the similarity metric $P$. The joint layers will take as input the concatenation of the previous outputs on both branches of the network and keep the AlexNet structure for following layers. Table~\ref{tab1} shows a detailed description of the architecture layers.

\begin{table}[!h]
  \caption{Detailed description of the proposed architecture using AlexNet for one-shot recognition}
  \label{tab1}
  \def\arraystretch{1.2}
  \begin{tabular}{P{3.0cm}|P{2cm}|P{2cm}}
    \hline
    Type & Filters&Size/Stride\\
    \hline
    Convolutional 1& 64& 11x11/4\\
    Maxpool& & 3x3/2\\
    Joint layer 1& 64& 3x3/1\\
    Convolutional 2& 192& 5x5/1\\
    Maxpool& & 3x3/2\\
    Joint layer 2& 192& 5x5/1\\
    Convolutional 3& 384& 3x3/1\\
    Joint layer 3& 384& 3x3/1\\
    Convolutional 4& 256& 3x3/1\\
    Convolutional 5& 256& 3x3/1\\
    Maxpool& & 3x3/2\\
    Linear& 4096& \\
    Linear& 4096& \\
    Linear& 1&\\
    \hline
\end{tabular}
\end{table}

In contrast to the traditional contrastive loss function
this work uses the binary cross entropy (BCE) loss function for training the proposed architecture as defined in equation~\ref{BCE}. The loss function is defined to comply with the SNN architecture previously defined and the specified one-shot constraints, where, the model is trained to output a similarity measure that is small for matches and big for dissimilar mismatches.

\begin{equation}
\begin{split}
\ell(x^{(1)},x^{(2)}) = -\frac{1}{N}\sum_{i=1}^{N} y_i(x^{(1)},x^{(2)}) \cdot log(p(x^{(1)},x^{(2)}))\\
+ (1 - y_i(x^{(1)},x^{(2)}) \cdot log(1 - p(x^{(1)},x^{(2)})),
\end{split}
\label{BCE}
\end{equation}

where $y$ is the matching label between the inputs $x^{(1)},x^{(2)}$, which equals to 1 for matching input pairs and 0 otherwise. $p(x^{(1)},x^{(2)})$ is the predicted label and $N$ is the batch size.

In order to achieve the one-shot detection task the proposed model is trained with categories ignored during testing. For this purpose the training data loader will select a set of random queries for each epoch with one single query image per batch matched with a random target image that contains the query image 50\% of times. Each query and target image is then passed through the model and paired with the ground truth annotations to compute the loss function.

\subsection{Detection}
\label{detectsect}

The presented recognition approach is adapted to perform multi-box one-shot object detection given a query and target images containing a pattern of interest from a category unseen during training. The proposed Joint Neural Network is defined as a multi-box detector based on the Darknet19 architecture~\cite{YOLO1}, taking the query and target images $x^{(1)}$, $x^{(2)}$ as input and outputting a tensor $\hat{P}$ representing the locations where the object of interest may be present in $x^{(2)}$.

Following~\cite{YOLO2} we define a $S \times S$ grid over the target image $x^{(2)}$. Each grid cell predicts $B$ anchor window coordinates and confidence score $(x, y, w, h, conf)$; where, the output predictions are encoded as a $S \times S \times (B \times 5)$ tensor. In contrast with~\cite{YOLO2} the category to be detected in the target image $x^{(2)}$ is dictated by the input query image $x^{(1)}$, so the output tensor does not require additional position to define prediction classes. The confidence scores are trained as intersection over union predictor which will have a higher value when its respective anchor window correctly contains the object of interest.

\begin{figure}
\centering
\includegraphics[width=200pt]{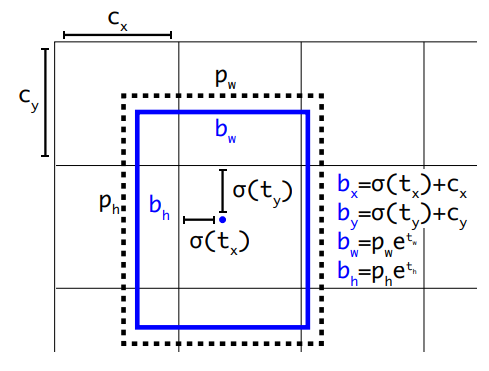}
\caption{Anchor window definition~\cite{YOLO2}. Where $(b_x, b_y)$ and $(b_w, b_h)$ define the center coordinates and size of the anchor window respectively}
\label{anchordef}
\end{figure}

The approach and notation for defining the anchor windows are inspired by~\cite{YOLO2} is used to define the anchor windows. For each anchor window in the $S \times S$ grid the model yields to five output values $t_x$, $t_y$, $t_w$, $t_h$, and $t_o$. The cell offsets from the top left corner of the image are defined as $(c_x, c_y)$ and the bounding box prior has width and height $p_w$, $p_h$. The resulting predictions are obtained using equations~\ref{bxeq} - \ref{confeq} as illustrated in Figure~\ref{anchordef}.

\begin{equation}
b_x = \sigma(t_x) + c_x
\label{bxeq}
\end{equation}
\begin{equation}
b_y = \sigma(t_y) + c_y
\label{byeq}
\end{equation}
\begin{equation}
b_w = p_we^{t_w}
\label{bweq}
\end{equation}
\begin{equation}
b_h = p_he^{t_h}
\label{bheq}
\end{equation}
\begin{equation}
conf = \sigma(t_o)
\label{confeq}
\end{equation}

\begin{figure*}
\centering
\includegraphics[width=0.90\textwidth]{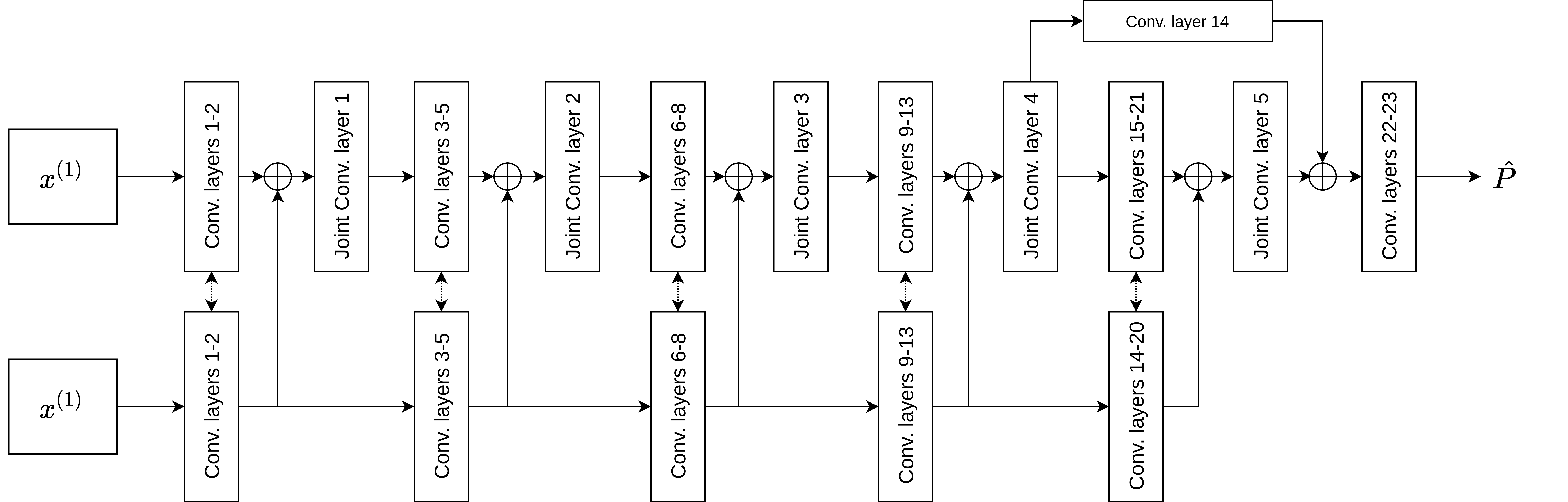}
\caption{JNN DarkNet19-based detection architecture.}
\label{arch}
\end{figure*}

The proposed model is essentially based on Darknet19~\cite{YOLO2}, adapted to take two inputs on separated branches. Both branches share weights and are connected by convolutional layers (\textit{joint layers}) at several points on the pipeline, ending in a single convolutional layer encoding the $S \times S \times (B \times 5)$ predictions tensor; where $S=14$ for the proposed architecture. Figure~\ref{arch} shows the base network structure for illustrative purposes in accordance with Table~\ref{arch}. The architecture used for experiments slightly differs from Figure \ref{arch} following the results presented in section \ref{ablation_section}.

\begin{table}[!b]
\centering
\caption{Detailed description of the proposed DarkNet19-based architecture.}\label{tab2}
\def\arraystretch{1.2}
\begin{tabular}{P{2.2cm}|P{1.00cm}|P{1.50cm}|P{1.50cm}}
Type & Filters & Size/Stride & Output \\
\hline
Convolutional 1 & 32 & 3x3 & 224x224\\
Maxpool &   & 2x2 / 2 & 112x112\\
Convolutional 2 & 64 & 3x3 & 112x112\\
Joint conv. 1 & 64 & 3x3 & 112x112\\
Convolutional 3 & 128 & 3x3 & 112x112\\
Convolutional 4 & 64 & 1x1 & 112x112\\
Convolutional 5 & 128 & 3x3 & 112x112\\
Maxpool &   & 2x2 / 2 & 56x56\\
Joint conv. 2 & 128 & 3x3 & 56x56\\
Convolutional 6 & 256 & 3x3 & 56x56\\
Convolutional 7 & 128 & 1x1 & 56x56\\
Convolutional 8 & 256 & 3x3 & 56x56\\
Maxpool &   & 2x2 / 2 & 28x28\\
Joint conv. 3 & 256 & 3x3 & 28x28\\
Convolutional 9 & 512 & 3x3 & 28x28\\
Convolutional 10 & 256 & 1x1 & 28x28\\
Convolutional 11 & 512 & 3x3 & 28x28\\
Convolutional 12 & 256 & 1x1 & 28x28\\
Convolutional 13 & 512 & 3x3 & 28x28\\
Joint conv. 4 & 512 & 1x1 & 28x28\\
Convolutional 14 & 64 & 1x1 & 28x28\\
Maxpool &   & 2x2 / 2 & 14x14\\
Convolutional 15 & 1024 & 3x3 & 14x14\\
Convolutional 16 & 512 & 1x1 & 14x14\\
Convolutional 17 & 1024 & 3x3 & 14x14\\
Convolutional 18 & 512 & 1x1 & 14x14\\
Convolutional 19 & 1024 & 3x3 & 14x14\\
Convolutional 20 & 1024 & 3x3 & 14x14\\
Convolutional 21 & 1024 & 3x3 & 14x14\\
Joint conv. 5 & 1024 & 3x3 & 14x14\\
Convolutional 22 & 1024 & 3x3 & 14x14\\
\hline
Convolutional 23 & $B * 5$ & 1x1 & 14x14\\
\hline
\end{tabular}
\end{table}

The input sizes for the target and query images are $448 \times 448$ and $224 \times 224$ respectively. By adapting the architecture to apply the first pooling layer on the target sister network only, the output sizes for the first 2 convolutional layers become even. This small adaptation makes the concatenation operation and further convolutional layers computation straightforward. In addition,the called joint convolutional layers in Figure~\ref{arch} are designed to take the concatenation of the previous convolutional layer outputs for each sister network as input and accommodate its output following the DarkNet19 architecture~\cite{YOLO2} in the same way as the residual residual layer.Finally, the last convolutional layer outputs a $14 \times 14$ tensor composed of $B \times 5$ filters that will encode the model predictions.

The loss function $\ell$ used for this work is defined as the weighted sum of losses for localisation and confidence components. The first term being the location loss $\ell_{loc}$, deals with how close the predicted box location is to the ground truth annotation; then, the confidence loss term $\ell_{conf}$ represents how certain the model is that the pattern of interest is located inside a specific anchor box.

\begin{equation}
\begin{split}
\ell = \ell_{loc} + \ell_{conf},
\end{split}
\label{loss}
\end{equation}

\begin{equation}
\begin{split}
\ell_{loc} = \lambda_{coord}\sum_{i=1}^{S^{2}} \sum_{j=1}^{B} \mathbbm{1}_{ij}^{obj}  (y - y_{pred})^2,
\end{split}
\label{locloss}
\end{equation}

\begin{equation}
\begin{split}
\ell_{conf} = \lambda_{obj}\sum_{i=1}^{S^{2}} \sum_{j=1}^{B} \Big[\mathbbm{1}_{ij}^{obj} c\cdot log(c_{pred})+\\
(1-c)\cdot log(1-c_{pred})\Big] + \\
\lambda_{noobj}\sum_{i=1}^{S^{2}} \sum_{j=1}^{B} \Big[\mathbbm{1}_{ij}^{noobj} c\cdot log(c_{pred})+\\
(1-c)\cdot log(1-c_{pred})\Big],
\end{split}
\label{confloss}
\end{equation}

\begin{figure*}[!h]
\centering
\includegraphics[width=380pt]{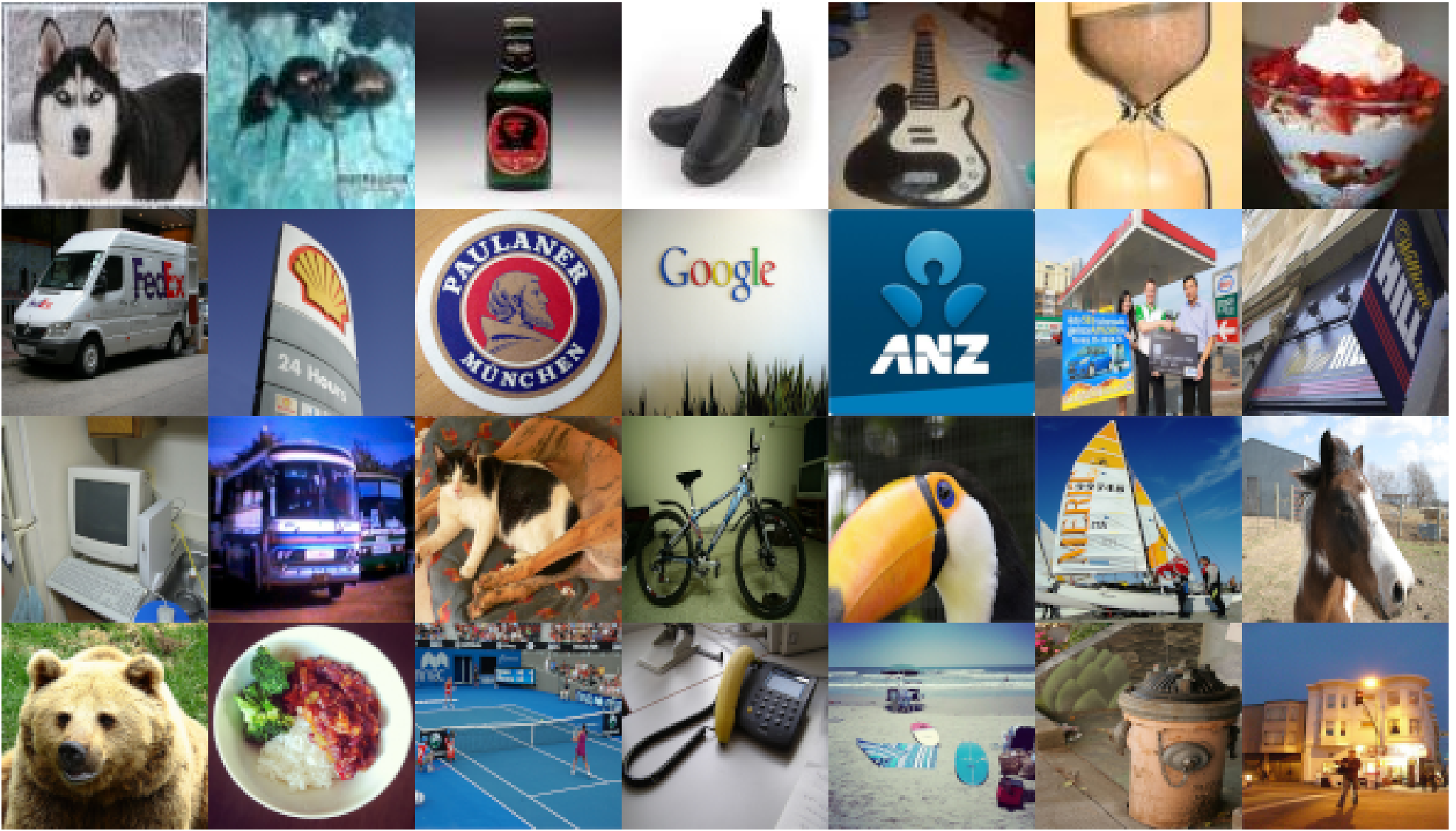}
\caption{Subset sample from the MiniImageNet (row 1), QMUL-OpenLogo (row 2), Pascal VOC (row 3), and COCO (row 4) datasets.} \label{datasamples}
\end{figure*}

where $S$ is the size of the grid; $B$ is the number of anchors defined per grid cell; $y$ and $y_{pred}$ are the ground truth and prediction tensors containing the $b_x$, $b_y$, $b_w$, $b_h$ values; $c$ and $c_{pred}$ are the ground truth and predicted confidences ($conf$); $\mathbbm{1}_{ij}^{obj}$ is equals to 1 where the anchor window $j$ at cell $i$ has an Intersection of Union $IoU>0.5$ with the target pattern and 0 otherwise; $\mathbbm{1}_{ij}^{noobj}$ is $\mathbbm{1}_{ij}^{obj}$ complement; and $\lambda_{coord}$, $\lambda_{obj}$, and $\lambda_{noobj}$ are weights to deal with the relevance of each term in the loss function. In contrast to~\cite{YOLO1} we use the Binary Cross Entropy function for the confidence terms along with the $\lambda_{obj}$, and $\lambda_{noobj}$ weights to deal with the learning priority when keeping the true positives or discarding false positives. The data loading for learning follows the same protocol as in recognition; where the target has a 50\% chance of containing the query pattern.

\section{Experiments}

This section presents the performance evaluation of the proposed approach initially for the recognition task using the AlexNet adaptation of the joint neural network architecture and afterwards using the Darknet variation for object detection. The experiments are built over three different datasets: The QMUL-OpenLogo, MiniImagenet, and pascalVOC datasets.

\textbf{QMUL-OpenLogo Dataset:} The OpenLogo dataset introduced in~\cite{paper24} is presented as a challenging benchmark to evaluate logo detection methods under realistic conditions. The dataset contains approximately 27,000 images from more than 300 logo classes captured from different sources. Figure~\ref{datasamples} (row 2) presents a small sample set that depicts the variance between the captures and the classes in the dataset.

The OpenLogo dataset exploits existing logo detection datasets; i.e., the dataset contains logo images from several different existing datasets and heterogeneous sources in an attempt to reproduce realistic scenarios~\cite{paper24}. These datasets are: FlickrLogos-27~\cite{52}, FlickrLogos-32~\cite{53}, Logo32plus~\cite{54}, BelgaLogos~\cite{55}, WebLogo-2M~\cite{56}, Logo-In-The-Wild~\cite{57}, SportsLogo~\cite{58}. To the best of our knowledge, the OpenLogo dataset has not been used by any published work in the context of one-shot recognition or detection.

The dataset is split into 210 categories for training and 125 categories for testing. For recongition tests both input images are cropped, while only the query image is cropped for detection experiments. Again, there is no category overlap between the training and testing sub-sets which guarantees that the embedded CNN has no knowledge of any of the tested query inputs and complies with the one-shot constraints. The categories on the QMUL-OpenLogo dataset are unbalanced, with some of the logos having as little data as 2 instances. Given that the proposed architecture works with pairs of images it is possible to make enough pairs testing purposes while keeping data balance. Therefore, approximately 20,000 input pairs are formed from the 125 testing categories available with a 50/50 split on matches and mismatches for the unseen testing categories.

\textbf{MiniImageNet Dataset:} The MiniImageNet dataset is derived from the ImageNet~\cite{44.1}, first proposed by~\cite{47}. MiniImageNet consists of 60,000 $84 \times 84$ images representing 100 classes with 600 instances each. The exact 600 images that comprise each of the 100 classes in miniImageNet can be found at \url{https://goo.gl/e3orz6} and the images can be directly downloaded from \url{https://github.com/yaoyao-liu/mini-imagenet-tools}. The miniImageNet dataset splits 80 classes for training and 20 for testing. Again, the training and testing classes do not overlap following the one-shot problem constraints. Figure~\ref{datasamples} presents (row 1) a set sample images from the miniImageNet dataset.

\textbf{Pascal VOC Dataset:} Pascal VOC is a collection of datasets for object detection \cite{VOC} used for the Pascal Visual Object Classes challenge. The most commonly used combination for benchmarking traditional object detection approaches is using the VOC 2007 trainval and 2012 trainval sets for training and the VOC 2007 test set for validation. These datasets contain 10.000 and 22.500 images respectively, and each image is annotated with one or several labels, corresponding to 20 object categories. Figure~\ref{datasamples} (row 3) presents a set of sample images from the VOC dataset.

\textbf{COCO Dataset:} COCO is a dataset that places the question of object recognition in the context of the broader question of scene understanding. The dataset is built from images of complex everyday scenes containing common objects in their natural context. The COCO dataset contains over a hundred thousand annotated images over 80 categories. The object categories in the data include animals, transport vehicles and furniture \cite{COCO}.

\subsection{Recognition results}
\label{recogressec}

In order to evaluate the performance of the proposed approach for the one-shot recognition task results are presented using true positive rates, false positive rates, accuracy, precision, and the ROC curve. In this subsection the proposed JNN based recognition approach is compared with the traditional SNN approach using the AlexNet architecture backbone over the OpenLogo dataset. Additionaly, the proposed approach is compared with few-shot learning methods \cite{46fs1}, \cite{46fs2}, \cite{46fs3}, \cite{46fs4}, \cite{46fs5} over the MiniImageNet dataset.

Given that different thresholds over the learned similarity metric $P$ lead to differences in the performance of the binary classifier, a set of 20 thresholds is used over the similarity metric within the range of responses returned by the Siamese neural network in order to compute the ROC curve. The threshold with the best accuracy is latter used for comparison purposes.

The proposed AlexNet variation of the architecture was trained 200 epochs with a $0.005$ learning rate. Each training batch contained 64 pairs of images randomly selected from the training set. The training process was around 9 hours to complete on a GeForce GTX 1080. The proposed approach as well as the Siamese AlexNet architecture used for experiments are implemented using the pythorch library.

\begin{figure}
\centering
\includegraphics[width=240pt]{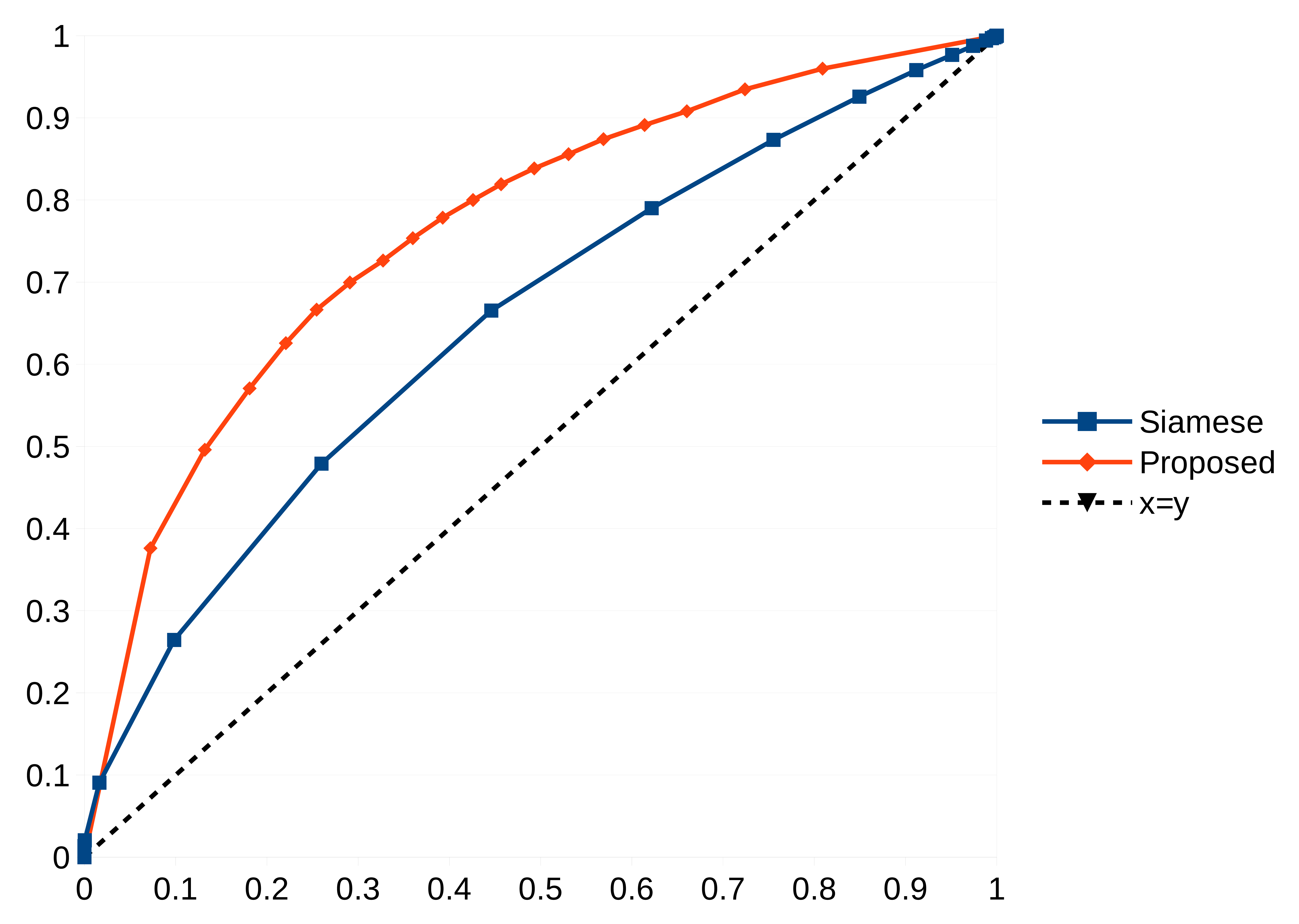}
\caption{$ROC$ curve showing the performance of the proposed approach and the standard SNN approach on the QMUL-OpenLogo dataset.} \label{logoROC}
\end{figure}

Figure~\ref{logoROC} shows the performance of the proposed JNN for logo recognition in terms of the $ROC$ curve obtained by thresholding the similarity metric $S$. For each one of the models, the similarity metric is thresholded in the range of responses of the model to compute and plot the true positive ($TPR$) and false positive ($FPR$) rates. The accuracy, precision, and area under the ROC curve ($AUC$), reported for each method is measured for the threshold with the highest accuracy. 

Table~\ref{tab3} shows the result scores for the approaches in Figure~\ref{logoROC}. The proposed JNN shows a significant improvement over the AUC, FPR and accuracy scores when compared with the standard SNN AlexNet approach ($AUC=0.76$, $FPR=0.25$, $acc=0.70$).

\begin{table}[h!]
\centering
\caption{Performance metrics for the proposed approach and SNN AlexNet.}\label{tab3}
\def\arraystretch{1.2}
\begin{tabular}{lccccc}
\hline
Method & TPR & FPR & acc & Pr & AUC \\
\hline
SNN & 0.67 & 0.44 & 0.60 & 0.59 & 0.65\\
\textbf{JNN} & 0.66 & \textbf{0.25} & \textbf{0.70} & \textbf{0.72} & \textbf{0.76}\\
\hline
\end{tabular}
\end{table}

Table~\ref{soaMiniTab} shows the accuracy scores for state-of-the-art one-shot learning approaches assessed in miniImageNet for object recognition. The proposed approach outperforms these methods without any fine-tuning or learning associated to the novel tested categories. This means that, compared to few-shot based learning, the presented JNN can fit a wider range of application scenarios, with reasonably accurate prediction outcomes at a significant smaller cost.

\begin{table}[h!]
\centering
\caption{Accuracy measured for the proposed approach compared to state-of-art approaches for one-shot recognition on the miniImageNet dataset.}\label{soaMiniTab}
\def\arraystretch{1.5}
\begin{tabular}{lc}
\hline
Method & accuracy (\%) \\
\hline
MALM (2017) & 48.70 \\
ProtoNets (2017) & 49.42 \\
SNAIL (2018) & 55.71 \\
TADAM (2019) & 58.50 \\
MLT (2019) & 61.20 \\
\textbf{JNN} & \textbf{61.41} \\
\hline
\end{tabular}
\end{table}

\begin{figure*}
\centering 
    \begin{subfigure}{0.3\textwidth}
    \centering
        \begin{subfigure}{0.4\textwidth}
            \centering \includegraphics[width=0.9\textwidth,height=2.0cm]{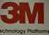}
        \end{subfigure}
        \begin{subfigure}{0.4\textwidth}
            \centering \includegraphics[width=0.9\textwidth,height=2.0cm]{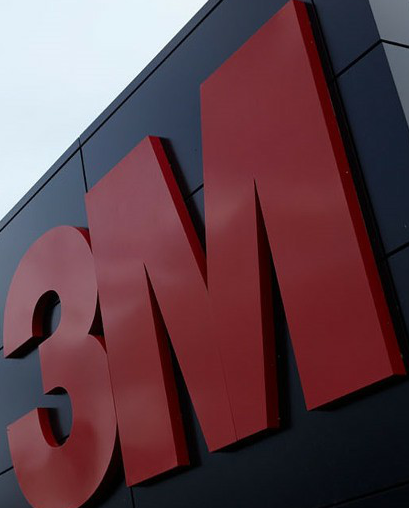}
        \end{subfigure}
        \caption{$P=0.63$}
    \end{subfigure}
\vrule
    \begin{subfigure}{0.3\textwidth}
    \centering
        \begin{subfigure}{0.4\textwidth}
            \centering
            \includegraphics[width=0.9\textwidth,height=2.0cm]{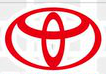}
        \end{subfigure}
        \begin{subfigure}{0.4\textwidth}
            \centering
            \includegraphics[width=0.9\textwidth,height=2.0cm]{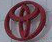}
        \end{subfigure}
    \caption{$P=0.53$}
    \end{subfigure}
\vrule
    \begin{subfigure}{0.3\textwidth}
    \centering
        \begin{subfigure}{0.4\textwidth}
            \centering \includegraphics[width=0.9\textwidth,height=2.0cm]{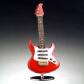}
        \end{subfigure}
        \begin{subfigure}{0.4\textwidth}
            \centering \includegraphics[width=0.9\textwidth,height=2.0cm]{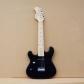}
        \end{subfigure}
    \caption{$P=0.62$}
    \end{subfigure}

    \begin{subfigure}{0.3\textwidth}
    \centering
        \begin{subfigure}{0.4\textwidth}
            \centering \includegraphics[width=0.9\textwidth,height=2.0cm]{3m1.png}
        \end{subfigure}
        \begin{subfigure}{0.4\textwidth}
            \centering \includegraphics[width=0.9\textwidth,height=2.0cm]{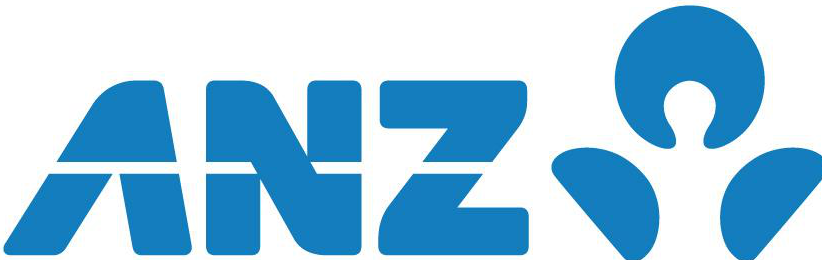}
        \end{subfigure}
        \caption{$P=0.13$}
    \end{subfigure}
\vrule
    \begin{subfigure}{0.3\textwidth}
    \centering
        \begin{subfigure}{0.4\textwidth}
            \centering
            \includegraphics[width=0.9\textwidth,height=2.0cm]{toyota22.png}
        \end{subfigure}
        \begin{subfigure}{0.4\textwidth}
            \centering
            \includegraphics[width=0.9\textwidth,height=2.0cm]{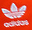}
        \end{subfigure}
    \caption{$P=0.15$}
    \end{subfigure}
\vrule
    \begin{subfigure}{0.3\textwidth}
    \centering
        \begin{subfigure}{0.4\textwidth}
            \centering
            \includegraphics[width=0.9\textwidth,height=2.0cm]{n0327201000000011.jpg}
        \end{subfigure}
        \begin{subfigure}{0.4\textwidth}
            \centering
            \includegraphics[width=0.9\textwidth,height=2.0cm]{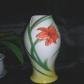}
        \end{subfigure}
    \caption{$P=0.23$}
    \end{subfigure}

\caption{One-shot recognition results for arbitrarily selected pairs of images from the QMUL-OpenLogo and MiniImageNet datasets.}
\label{recogres}
\end{figure*}

Figure~\ref{recogres} depicts the predicted value $P$ for a few arbitrarily selected patterns from the OpenLogo and MiniImageNet datasets unseen during training.

\begin{table*}[h!]
\centering
\caption{ Comparison of different one-shot detection methods on VOC in AP (\%). Note that the SiamFC  \cite{SiamFC}, SiamRPN \cite{SiamRPN}, CompNet \cite{CompNet}, CoAE \cite{coAE}, and M-FCOS+SARM \cite{MFCOS} approaches use all classes in their ImageNet pre-trained backbones.}
\label{soaVOCpre}
\begin{tabular}{p{1.1cm}|p{0.3cm}p{0.3cm}p{0.3cm}p{0.3cm}p{0.3cm}p{0.3cm}p{0.3cm}p{0.5cm}p{0.2cm}p{0.3cm}p{0.3cm}p{0.3cm}p{0.3cm}p{0.3cm}p{0.3cm}p{0.3cm}p{0.45cm}|p{0.25cm}p{0.35cm}p{0.3cm}p{0.3cm}p{0.45cm}}
\hline
Method & \multicolumn{17}{c}{Seen classes} & \multicolumn{5}{c}{Unseen classes}\\ 
& plant & sofa & tv & car & bottle & boat & chair & person & bus & train & horse & bike & dog & bird & mbike & table & mAP & cow & sheep & cat & aero & mAP \\
\hline
SiamFC & 3.2 & 22.8 & 5.0 & 16.7 & 0.5 & 8.1 & 1.2 & 4.2 & 22.2 & 22.6 & 27.8 & 35.4 & 14.2 & 25.8 & 11.7 & 19.7 & 15.1 & 6.8 & 2.28 & 31.6 & 12.4 & 13.3 \\
SiamRPN &  1.9 & 15.7 & 4.5 & 12.8 & 1.0 & 1.1 & 6.1 & 8.7 & 7.9 & 6.9 & 5.1 & 17.4 & 17.8 & 20.5 & 7.2 & 18.5 & 9.6 & 15.9 & 15.7 & 21.7 & 3.5 & 14.2 \\
CompNet & 28.4 & 41.5 & 65.0 & 66.4 & 37.1 & 49.8 & 16.2 & 31.7 & 69.7 & 73.1 & 39.8 & 75.6 & 71.6 & 61.4 & 52.3 & 63.4 & 52.7 & 75.3 & 60.0 & 47.9 & 25.3 & 52.1 \\
CoAE & 30.0 & 54.9 & 64.1 & 66.7 & 40.1 & 54.1 & 14.7 & 60.9 & 77.5 & 78.3 & 46.2 & 77.9 & 73.2 & 80.5 & 70.8 & 72.4 & 60.1 & 83.9 & 67.1 & 75.6 & 46.2 & 68.2 \\
M-FCOS & 33.7 & 58.2 & 67.5 & 72.7 & 40.8 & 48.2 & 20.1 & 55.4 & 78.2 & 79.0 & 48.8 & 76.2 & 74.6 & 81.3 & 71.6 & 72.0 & \textbf{61.1} & 74.3 & 68.5 & 81.0 & 52.4 & \textbf{69.1} \\
\textbf{JNN} & 4.5 & 44.0 & 38.3 & 51.5 & 2.9 & 43.76 & 13.1 & 39.3 & 45.2 & 57.8 & 72.6 & 58.0 & 75.5 & 64.2 & 71.14 & 12.7 & 43.4 & 64.7 & 51.0 & 65.2 & 43.5 & 56.1 \\
\hline
\end{tabular}
\end{table*}

\begin{table*}[h!]
\centering
\caption{One-shot object detection performance comparison training on the COCO dataset and testing using the Pascal VOC dataset.}
\label{COCO2VOCtable}

\begin{tabular}{p{1.1cm}|p{0.3cm}p{0.3cm}p{0.3cm}p{0.3cm}p{0.35cm}p{0.3cm}p{0.3cm}p{0.5cm}p{0.2cm}p{0.3cm}p{0.3cm}p{0.3cm}p{0.3cm}p{0.3cm}p{0.35cm}p{0.3cm}p{0.3cm}p{0.35cm}p{0.3cm}p{0.38cm}|p{0.43cm}}

\hline
Method & plant & sofa & tv & car & bottle & boat & chair & person & bus & train & horse & bike & dog & bird & mbike & table & cow & sheep & cat & aero & mAP \\
\hline
M-FCOS & 10.3 & 79.1 & 54.2 & 64.0 & 18.4 & 19.6 & 26.3 & 9.6 & 67.5 & 55.4 & 67.7 & 26.8 & 73.7 & 50.8 & 26.4 & 11.9 & 84.8 & 75.1 & 69.5 & 28.4 & 45.9 \\
\textbf{JNN} & 9.5 & 69.3 & 49.8 & 60.3 & 7.2 & 29.1 & 10.1 & 6.7 & 60.3 & 57.2 & 58.5 & 45.3 & 62.6 & 45.6 & 74.8 & 29.0 & 70.4 & 55.4 & 54.4 & 88.1 & \textbf{47.1} \\
\hline
\end{tabular}
\end{table*}

\subsection{Detection results}
\label{detectressect}

The proposed JNN based detection approach is tested under four experimental settings using the QMUL-OpenLogo, VOC, and COCO datasets. First, we follow the experiment proposed in \cite{coAE} by training the proposed model in 16 of the 20 classes of the VOC dataset and testing on the additional 4 classes. The model is trained using the VOC 2007 and 2012 train\&val set and evaluated using the VOC 2007 test set. Second, we use the experimental approach presented in \cite{XiangLi}; where the proposed model is trained in the COCO dataset and tested in the VOC excluding the overlapping classes during training. Third, following \cite{SiamRPN} the model is trained and tested using the COCO dataset using four different splits for the training and testing classes. Finally, the model is tested on the QMUL-OpenLogo dataset using a 60/40 split for training and testing classes selected arbitrarily and complying with the one-shot constraints. 

The proposed JNN is trained using the SGD optimizer with a learning rate equals to 0.0001, a 0.9 momentum for 160 epochs, with a 16 batch size, over a single GeForce GTX 1080 GPU. All the experiments are performed using an $IoU$ threshold of 0.5.

Table \ref{soaVOCpre} shows the results for the first experimental setting, along with the train and test class splits for the VOC dataset as in \cite{coAE}. Under this experimental settings, results show a 13\% mAP difference to the top performing approach \cite{XiangLi}. However, it should be noted that in contrast to the results presented from the previous approaches we do not pre-train our model. The results table presented in \cite{XiangLi} use the pretrained models on the full ImageNet dataset where the test classes may be present, meaning that the one-shot constraint is not in strict use.

The authors in \cite{XiangLi} go further, experimenting with one-shot object detection using the COCO and VOC datasets. Following their approach, the second experimental setting presented in this paper for one-shot object detection trains the proposed model on the COCO dataset excluding the VOC classes used for testing. Table \ref{COCO2VOCtable} shows the results compared to the M-FCOS approach \cite{XiangLi}. Results show a mAP improvement using the JNN based detection JNN based detection approach presented in this paper when compared to the top performing approach \cite{XiangLi} in Table \ref{soaVOCpre}.

In addition, the performance of the JNN is also compared to the state-of-the-art approaches using the experimental approach presented in \cite{SiamRPN} and \cite{coAE}. In this case, four different splits of training and testing categories are defined over the COCO dataset. The fully detailed category splits can be found in the first appendix of \cite{SiamRPN}. Each training split will contain 60 categories leaving 20 for testing. Table \ref{COCO2Splittable} shows the results of the proposed approach compared to the SiamRPN \cite{SiamRPN} and coAcoE \cite{coAE} approaches. Under this experimental setting our approach is down 8.5\% mAP when compared with the CoAE approach \cite{coAE}.

\begin{table}
\centering
\caption{Evaluation on COCO with respect to mAP score (\%) for unseen classes.}
\label{COCO2Splittable}
\begin{tabular}{p{2.0cm}|c c c c|c}
\hline
split & 1 & 2 & 3 & 4 & Average\\
\hline
SiamRPN & 15.3 & 17.6 & 17.4 & 17.0 & 16.8 \\
coAcoE &  23.4 & 23.6 & 20.5 & 20.4 & \textbf{22.0} \\
\textbf{JNN} & 12.9 & 16.5 & 10.0 & 14.4 & 13.5 \\
\hline
\end{tabular}
\end{table}

\begin{figure*}
\centering 

    \begin{subfigure}{0.3\textwidth}
    \centering
        \begin{subfigure}[t]{0.2\textwidth}
            \centering
            \includegraphics[width=0.9\textwidth,height=1.0cm]{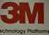}
        \end{subfigure}
        \begin{subfigure}{0.78\textwidth}
            \centering
            \includegraphics[width=0.9\textwidth,height=3.5cm]{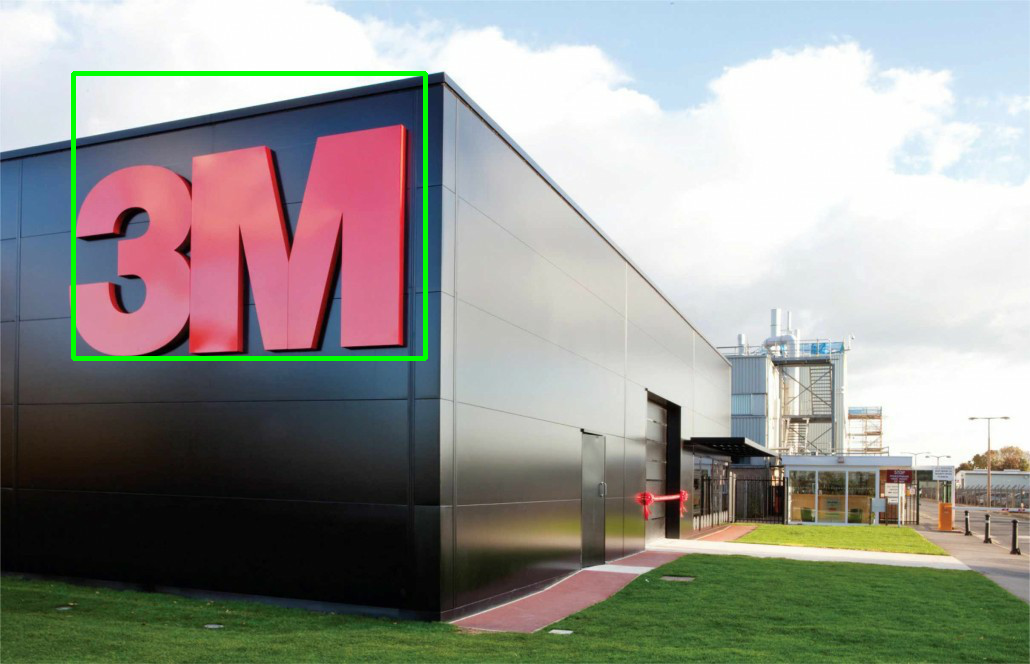}
        \end{subfigure}
    \end{subfigure}
\vrule
    \begin{subfigure}{0.3\textwidth}
    \centering
        \begin{subfigure}[t]{0.2\textwidth}
            \centering
            \includegraphics[width=0.9\textwidth,height=1.0cm]{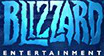}
        \end{subfigure}
        \begin{subfigure}{0.78\textwidth}
            \centering
            \includegraphics[width=0.9\textwidth,height=3.5cm]{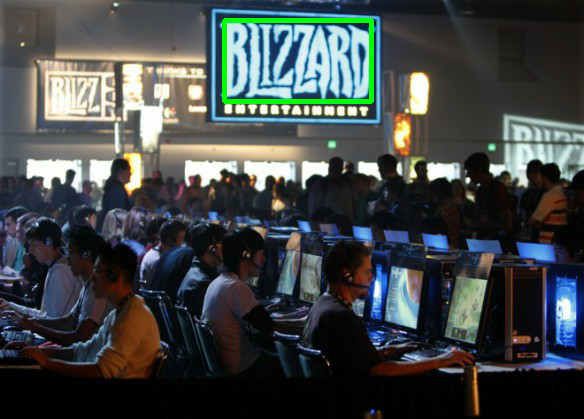}
        \end{subfigure}
    \end{subfigure}
\vrule
    \begin{subfigure}{0.3\textwidth}
    \centering
        \begin{subfigure}[t]{0.2\textwidth}
            \centering
            \includegraphics[width=0.9\textwidth,height=1.0cm]{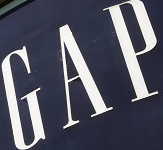}
        \end{subfigure}
        \begin{subfigure}{0.78\textwidth}
            \centering
            \includegraphics[width=0.9\textwidth,height=3.5cm]{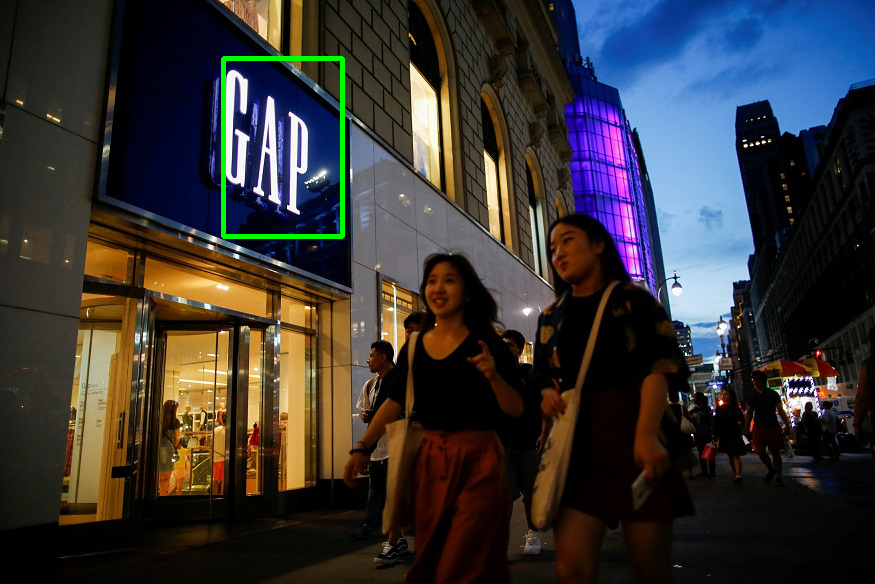}
        \end{subfigure}
    \end{subfigure}
    
    \begin{subfigure}{0.3\textwidth}
    \centering
        \begin{subfigure}[t]{0.2\textwidth}
            \centering
            \includegraphics[width=0.9\textwidth,height=1.0cm]{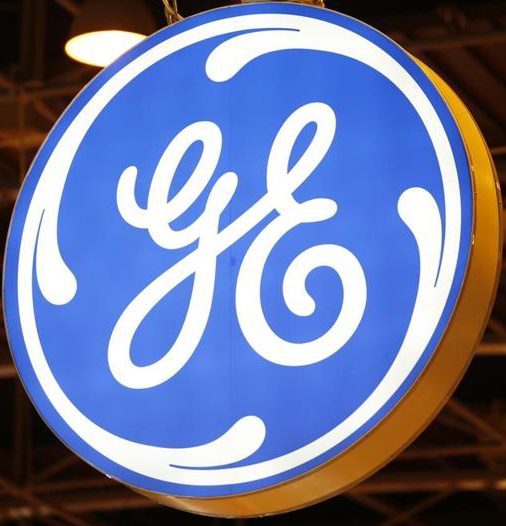}
        \end{subfigure}
        \begin{subfigure}{0.78\textwidth}
            \centering
            \includegraphics[width=0.9\textwidth,height=3.5cm]{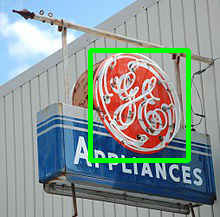}
        \end{subfigure}
    \end{subfigure}
\vrule
    \begin{subfigure}{0.3\textwidth}
    \centering
        \begin{subfigure}[t]{0.2\textwidth}
            \centering
            \includegraphics[width=0.9\textwidth,height=1.0cm]{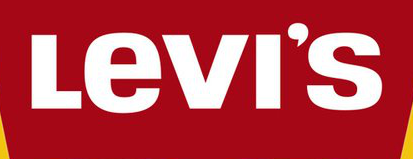}
        \end{subfigure}
        \begin{subfigure}{0.78\textwidth}
            \centering
            \includegraphics[width=0.9\textwidth,height=3.5cm]{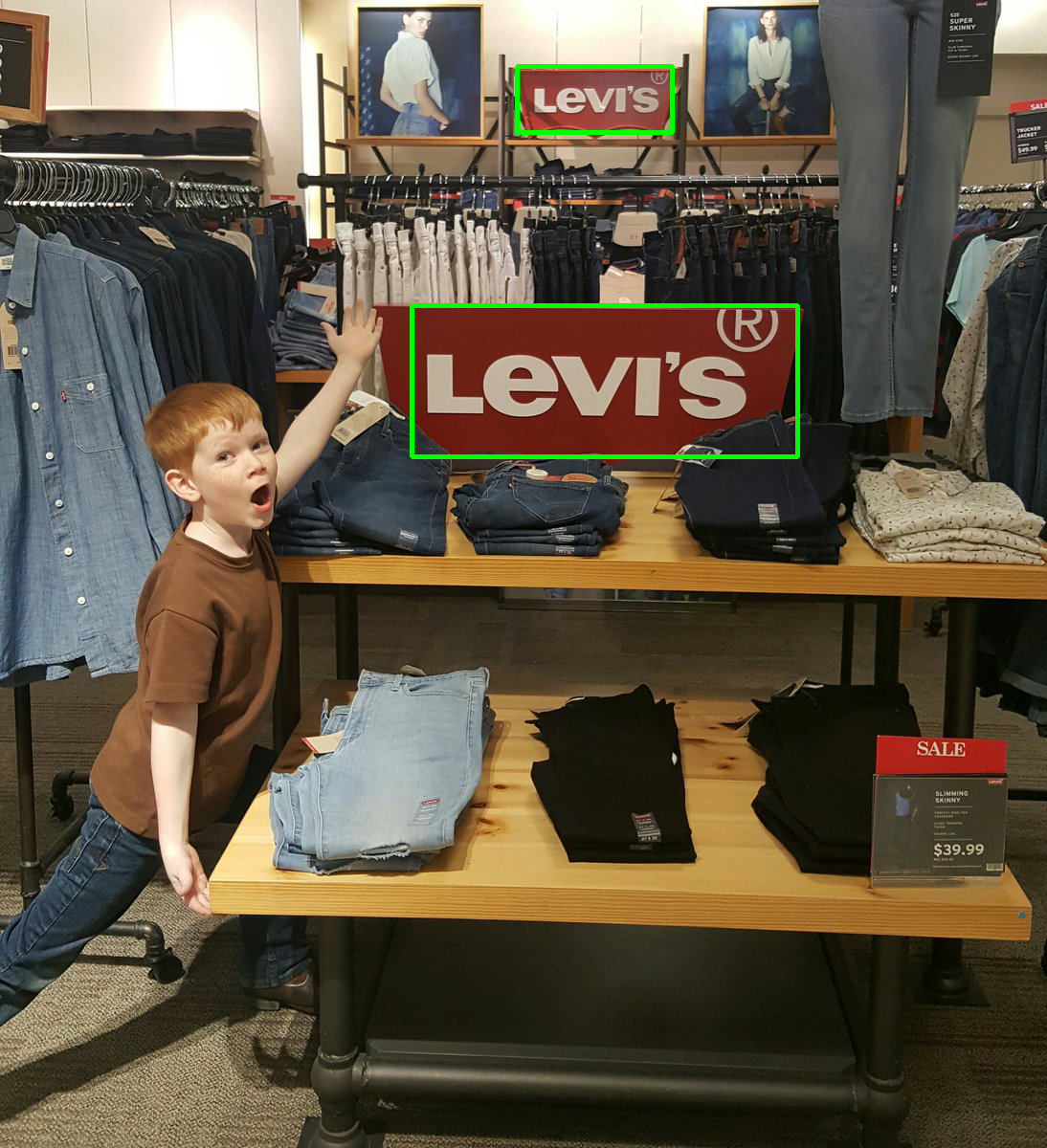}
        \end{subfigure}
    \end{subfigure}
\vrule
    \begin{subfigure}{0.3\textwidth}
    \centering
        \begin{subfigure}[t]{0.2\textwidth}
            \centering
            \includegraphics[width=0.9\textwidth,height=1.0cm]{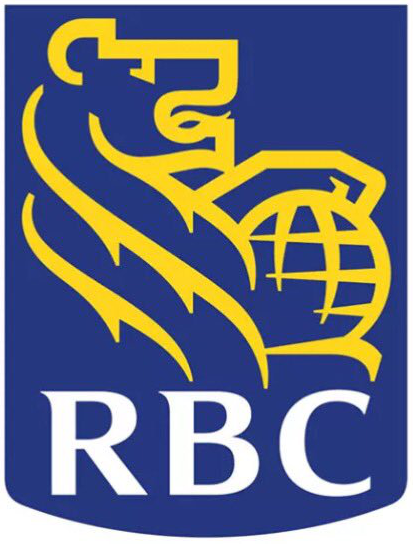}
        \end{subfigure}
        \begin{subfigure}{0.78\textwidth}
            \centering
            \includegraphics[width=0.9\textwidth,height=3.5cm]{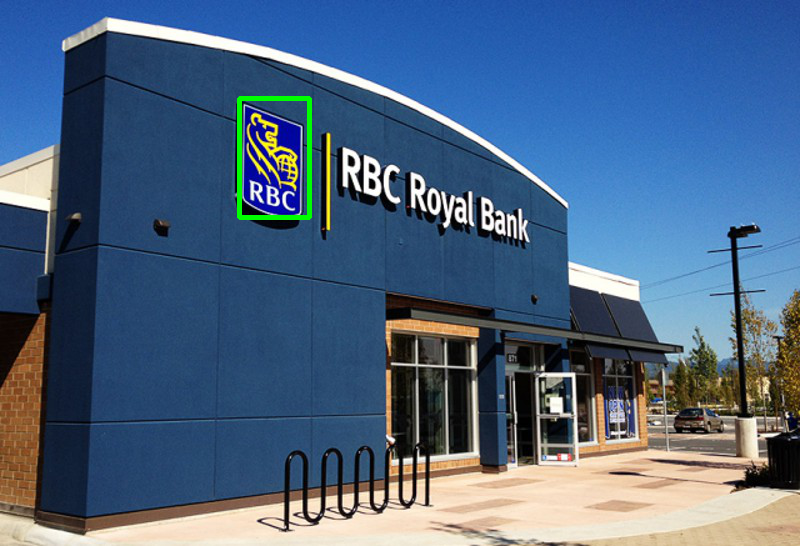}
        \end{subfigure}
    \end{subfigure}
        
    \begin{subfigure}{0.3\textwidth}
    \centering
        \begin{subfigure}[t]{0.2\textwidth}
            \centering
            \includegraphics[width=0.9\textwidth,height=1.0cm]{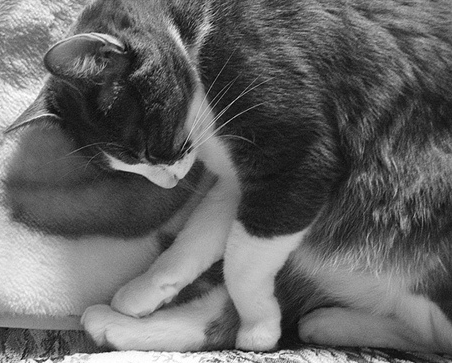}
        \end{subfigure}
        \begin{subfigure}{0.78\textwidth}
            \centering
            \includegraphics[width=0.9\textwidth,height=3.5cm]{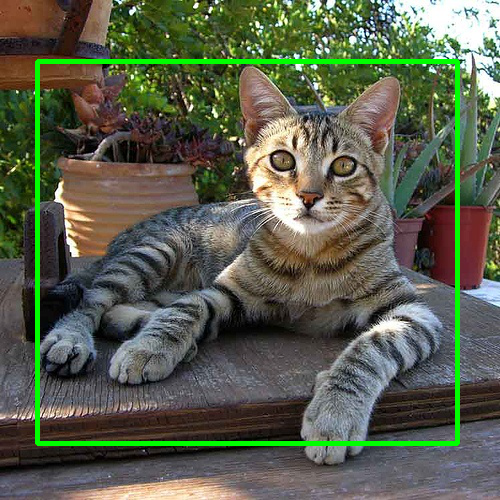}
        \end{subfigure}
    \end{subfigure}
\vrule
    \begin{subfigure}{0.3\textwidth}
    \centering
        \begin{subfigure}[t]{0.2\textwidth}
            \centering
            \includegraphics[width=0.9\textwidth,height=1.0cm]{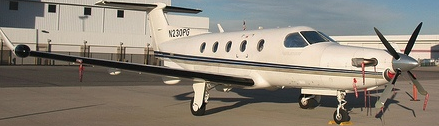}
        \end{subfigure}
        \begin{subfigure}{0.78\textwidth}
            \centering
            \includegraphics[width=0.9\textwidth,height=3.5cm]{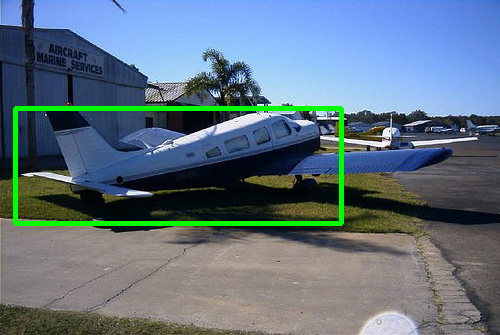}
        \end{subfigure}
    \end{subfigure}
\vrule
    \begin{subfigure}{0.3\textwidth}
    \centering
        \begin{subfigure}[t]{0.2\textwidth}
            \centering
            \includegraphics[width=0.9\textwidth,height=1.0cm]{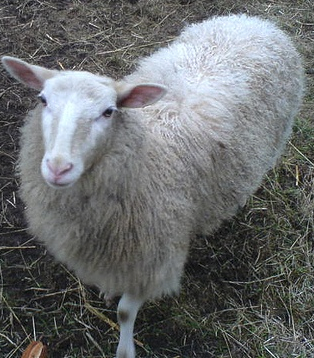}
        \end{subfigure}
        \begin{subfigure}{0.78\textwidth}
            \centering
            \includegraphics[width=0.9\textwidth,height=3.5cm]{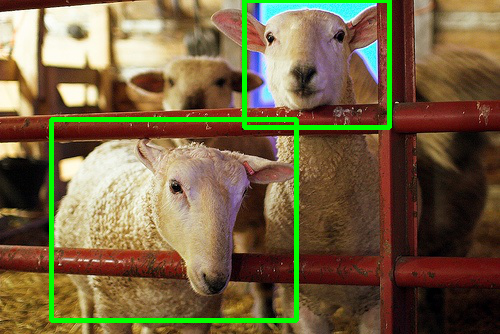}
        \end{subfigure}
    \end{subfigure}
    
    \begin{subfigure}{0.3\textwidth}
    \centering
        \begin{subfigure}[t]{0.2\textwidth}
            \centering
            \includegraphics[width=0.9\textwidth,height=1.0cm]{bear_query000000000285.jpg}
        \end{subfigure}
        \begin{subfigure}{0.78\textwidth}
            \centering
            \includegraphics[width=0.9\textwidth,height=3.5cm]{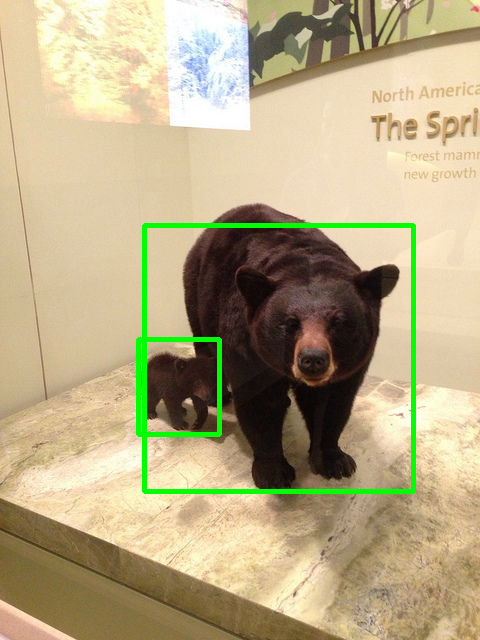}
        \end{subfigure}
    \end{subfigure}
\vrule
    \begin{subfigure}{0.3\textwidth}
    \centering
        \begin{subfigure}[t]{0.2\textwidth}
            \centering
            \includegraphics[width=0.9\textwidth,height=1.0cm]{bus_query000000001584.jpg}
        \end{subfigure}
        \begin{subfigure}{0.78\textwidth}
            \centering
            \includegraphics[width=0.9\textwidth,height=3.5cm]{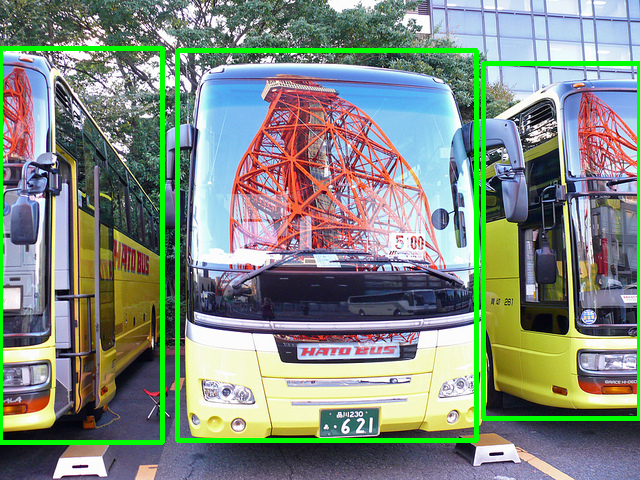}
        \end{subfigure}
    \end{subfigure}
\vrule
    \begin{subfigure}{0.3\textwidth}
    \centering
        \begin{subfigure}[t]{0.2\textwidth}
            \centering
            \includegraphics[width=0.9\textwidth,height=1.0cm]{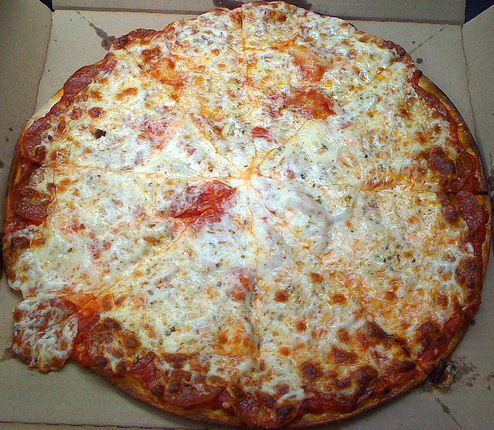}
        \end{subfigure}
        \begin{subfigure}{0.78\textwidth}
            \centering
            \includegraphics[width=0.9\textwidth,height=3.5cm]{pizza_target000000306700.png}
        \end{subfigure}
    \end{subfigure}
    
\caption{One-shot detection results results for arbitrarily selected query and target images from the QMUL-OpenLogo dataset.}
\label{detectres}
\end{figure*}

Finally, the proposed approach is assessed in the context of one-shot logo detection using the QMUL-OpenLogo dataset. Here, 210 categories are arbitrarily selected for training and 125 for testing in the same way a the recognition experiments in section \ref{recogressec}. Table \ref{mAPtab} shows the average precision (AP50) results for the top performing 15 unseen classes and the mean average precision (mAP) for the all 125 unseen classes used for testing. Figure~\ref{detectres} show some results of the proposed approach in action.

\begin{table}
\centering
\caption{AP detection results for the top 15 Open-Logo dataset classes and the mAP results for the whole dataset.}
\label{mAPtab}
\def\arraystretch{1.2}
\begin{tabular}{ll}
\hline
Class & mAP/AP \\
\hline
anz\_text               & 100.00\%\\
rbc                     & 98.86\%\\
blizzardentertainment   & 98.34\%\\
costco                  & 93.26\%\\
3m                      & 90.61\%\\
bosch\_text             & 90.00\%\\
gap                     & 89.47\%\\
lexus                   & 88.92\%\\
generalelectric         & 83.32\%\\
hp                      & 82.81\%\\
levis                   & 79.36\%\\
airhawk                 & 79.17\%\\
danone                  & 79.02\%\\
armitron                & 77.73\%\\
google                  & 77.66\%\\
\hline
all                     & \textbf{52.84}\%\\
\hline
\end{tabular}
\end{table}

\subsection{Ablation study}
\label{ablation_section}

This paper introduces Joint Neural Networks as an adaptable framework that can be used with any defined backbone architecture to perform the one-shot object recognition or detection tasks, where the joint layers are the key concept used build the Joint Neural Networks. This section explores how the use of joint layers on specific locations on the architecture may affect the performance of the proposed approach.

Figure \ref{arch} presented in section \ref{detectsect} shows the proposed convolutional and joint layers used to tackle the one-shot object detection task. In order to test the firstly proposed architecture, Table \ref{ablationTable} shows the performance of the proposed framework under different joint layer configurations. The test are carried out using the first experimental settings described in section \ref{detectressect} (Table \ref{soaVOCpre}).

\begin{table}
\centering
\caption{Ablation study for used joint layers in DarkNet19 architecture for detection}
\label{ablationTable}
\begin{tabular}{ccccccc}
\hline
JL1 & JL2 & JL3 & JL4 & JL5 & mAP\\
\hline
$\times$ &  &  &  &  &  53.3 \\
$\times$ & $\times$ &  &  &  &  53.6 \\
$\times$ & $\times$ & $\times$ &  &  &  54.9 \\
$\times$ & $\times$ & $\times$ & $\times$ &   & 55.2 \\
$\times$ & $\times$ & $\times$ & $\times$ & $\times$  & 36.8 \\
$\times$ &  &  & $\times$ &  &  52.9 \\
 & $\times$ &  & $\times$ &  &  53.9 \\
 &  & $\times$ & $\times$ &  &  47.9 \\
$\times$ & $\times$ &  & $\times$ &  & \textbf{56.1} \\
$\times$ &  & $\times$ & $\times$ &  & 53.7\\
\hline
\end{tabular}
\end{table}

Each Joint Layer ($JL$) in Figure \ref{arch} is depicted as a column in Table \ref{ablationTable}. The experiments explore the performance of the architecture when adding joint layers sequentialy as well as for groups of two and three layers. Results show that for the one-shot object detection task using the VOC dataset the combination of the joint layers 1, 2, and 4 in DarkNet19 architecture yields the best overall mAP performance (56.1\%).

\section{Conclusion}

This paper presents a novel convolutional neural network based framework named Joint neural networks to tackle the challenging tasks of one-shot object recognition and detection. Joint neural networks are inspired by Siamese neural networks and one-stage multi-box object detection approaches. The proposed joint layers are the fundamental concept supporting the data sharing between the branches of the presented Joint architectures. This approach is developed for one-shot object recognition and detection tasks where traditional deep learning approaches fall short due to the lack of training data available to build state-of-the-art models.

Results indicate that the proposed approach can compete with the top performing state-of-the-art recognition and detection approaches, and even outperform them under the right experimental settings by using the most basic building blocks of convolutional neural networks. The work presented in this paper leaves open room to experiment with new architectures and joint layer configurations to further improve the state-of-the-art performance for the one-shot object recognition and detection tasks.

\ifCLASSOPTIONcaptionsoff
  \newpage
\fi


\begin{thebibliography}{99}

\bibitem{paper1}
A. Jain, R. P. W. Duin, Jianchang Mao. ``Statistical pattern recognition: a review". \textit{IEEE Trans. Pattern Analysis and Machine Intelligence}, vol. 22, no. 1, pp. 4--37, Jan 2000. 

\bibitem{Jref1}
Z. Zhao, P. Zheng, S. Xu and X. Wu, ``Object Detection With Deep Learning: A Review", \textit{IEEE Trans. Neural Netw. Learn. Syst}, vol. 30, no. 11, pp. 3212-3232, Nov. 2019.

\bibitem{paper2}
A. Antoniou, A. Storkey, H. Edwards. ``Data augmentation generative adversarial networks", 2017. [Online] Available: \url{https://arxiv.org/abs/1711.04340}.

\bibitem{paper3}
A. Nguyen, J. Yosinski, J, Clune. ``Deep Neural Networks are Easily Fooled: High Confidence Predictions for Unrecognizable Images", December 2014. [Online] Available: \url{https://arxiv.org/abs/1412.1897}.

\bibitem{YOLO1}
Redmon, J., Divvala, S., Girshick, R., Farhadi, A. ``You only look once: Unified, real-time object detection", In \textit{Proc. CVPR}, Las Vegas, NV, USA 2016.

\bibitem{YOLO2}
J. Redmon and A. Farhadi, ``Yolo9000: better, faster, stronger", In \textit{Proc. CVPR}, 2017, pp. 6517–6525.

\bibitem{SIFT}
Lowe, David G. ``Distinctive Image Features from Scale-Invariant Keypoints", \textit{IJCV}, 60(2):91–110, November 2004.

\bibitem{SURF}
Bay, Herbert, Andreas Ess, Tinne Tuytelaars, and Luc Van Gool. ``Speeded-Up Robust Features (SURF)", \textit{Computer Vision and Image Understanding}, Volume 110, Issue 3, June 2008.

\bibitem{paper5}
G. Koch, R. Zemel, R. Salakhutdinov. ``Siamese neural networks for one-shot image recognition". \textit{ICML Deep Learning workshop}, 2015.

\bibitem{paper6}
L. Fe-Fei, R. Fergus, P. Perona. ``A Bayesian approach to unsupervised one-shot learning of object categories". In \textit{Proc. ICCV}, vol. 2, pp. 1134-1141, 2003.

\bibitem{paper7}
L. Fei-Fei, R. Fergus, P. Perona. ``One-shot learning of object categories".  \textit{IEEE Trans. Pattern Analysis and Machine Intelligence}, vol. 28, no.4, pp. 594–611, April 2006.

\bibitem{thesis37}
Nikolenko, S. I. ``Synthetic data for deep learning", 2019. [Online]  Available: \url{https://arxiv.org/abs/1909.11512}.

\bibitem{thesis36}
N. Tajbakhsh. ``Convolutional neural networks for medical image analysis: Full training or fine tuning?", \textit{IEEE Trans. Med. Imag}, vol. 35, no. 5, pp. 1299-1312, May 2016. 

\bibitem{Jref2}
L. Shao, F. Zhu and X. Li, ``Transfer Learning for Visual Categorization: A Survey", \textit{IEEE Trans. Neural Netw. Learn. Syst}, vol. 26, no. 5, pp. 1019-1034, May 2015.

\bibitem{thesis31lost}
R. Girshick, J. Donahue, T. Darrell, and J. Malik. ``Rich feature hierarchies for accurate object detection and semantic segmentation". \textit{Proc. CVPR}, 2014.

\bibitem{thesis32lost}
M. D. Zeiler and R. Fergus. ``Visualizing and understanding convolutional networks". In \textit{Proc. CVPR}, 2014.

\bibitem{thesis32lost2}
G. E. Hinton and R. R. Salakhutdinov. ``Reducing the dimensionality of data with neural networks". Science, 313(5786):504–507, 2006.

\bibitem{thesis32lost3}
Y. Wang, D. Ramanan, and M. Hebert, ``Growing a brain: Fine-tuning by increasing model capacity", in \textit{IEEE Comput. Soc. Conf. Comput. Vis. Pattern Recognit}, Jul. 2017, pp. 3029–3038.

\bibitem{46fs1}
C. Finn, P. Abbeel, and S. Levine. ``Model-agnostic metalearning for fast adaptation of deep networks". In \textit{International Conference on Machine Learning}, 2017

\bibitem{46fs2}
J. Snell, K. Swersky, and R. S. Zemel. ``Prototypical networks for few-shot learning". In \textit{Neural Information Processing Systems}, 2017. 

\bibitem{46fs3}
Nikhil Mishra, Mostafa Rohaninejad, Xi Chen, and Pieter Abbeel. ``A simple neural attentive meta-learner". In \textit{International Conference on Learning Representations}, 2018.

\bibitem{46fs4}
B. N. Oreshkin, P. Rodriguez, and A. Lacoste. ``TADAM: Task dependent adaptive metric for improved few-shot learning", May 2018. [Online] Available: \url{https://arxiv.org/abs/1805.10123}

\bibitem{46fs5}
Sun, Q., Liu, Y., Chua, T.S., Schiele, B. ``Meta-transfer learning for few-shot learning". In \textit{Proc. CVPR}, 2019.

\bibitem{Jref3}
H. -G. Jung and S. -W. Lee, ``Few-Shot Learning With Geometric Constraints", \textit{IEEE Trans. Neural Netw. Learn. Syst}, vol. 31, no. 11, pp. 4660-4672, Nov. 2020.

\bibitem{paper13}
E. Ahmed, M. Jones, T. K. Marks. ``An improved deep learning architecture for person re-identification". In \textit{Proc. CVPR}, 2015.

\bibitem{paper14}
Y. Taigman, M. Yang, M. Ranzato, L. Wolf: Deepface. ``Closing the gap to human-level performance in face verification". In \textit{Proc. CVPR}, 2014.

\bibitem{paperref24} 
Z. Zhang, H. Peng, and Q. Wang. ``Deeper and wider siamese networks for real-time visual tracking". In \textit{Proc. CVPR}, 2019.

\bibitem{paperref23} 
L. Bertinetto, J. Valmadre, J. F. Henriques, A. Vedaldi, and P. H. Torr. ``Fully-convolutional siamese networks for object tracking". In \textit{Proc. ECCV workshop}, 2016.

\bibitem{paperref26}
Bromley, J., Bentz, J. W., Bottou, L., Guyon, I., LeCun, Y., Moore, C., et al. ``Signature verification using a Siamese time delay neural network". \textit{Proc. Advances in Neural Information Processing Systems}, 6, pp. 737-744, 1994.

\bibitem{paperref27}
S. Chopra, R. Hadsell, and Y. LeCun. ``Learning a similiarty metric discriminatively, with application to face verification". In \textit{Proc. CVPR}, 2005.

\bibitem{paperref28}
M. Harandi, S. R. Kumar, and R. Nock. ``Siamese Networks: A Thing or Two to Know". 2017. [Online]  Available: \url{https://pdfs.semanticscholar.org/c03c/e09b419b6a15c1228e344a900d8c54bddc78.pdf}

\bibitem{coAE}
Hsieh, T.I., Lo, Y.C., Chen, H.T., Liu, T.L. ``One-Shot Object Detection with Co-Attention and Co-Excitation"". In \textit{Proc. NeurIPS}, 2019.

\bibitem{MFCOS}
Xiang Li, Lin Zhang, Yau Pun Chen, Yu-Wing Tai, Chi-Keung Tang. ``One-Shot Object Detection without Fine-Tuning", 2020. [Online]  Available: \url{https://arxiv.org/abs/2005.03819}.

\bibitem{SiamRPN}
Bo Li, Junjie Yan, Wei Wu, Zheng Zhu, and Xiaolin Hu. ``High performance visual tracking with siamese region proposal network". In \textit{Proc. CVPR}, 2018, Salt Lake City, UT, USA, June 18-22, 2018. 

\bibitem{SiamFC}
Miaobin Cen and Cheolkon Jung. ``Fully convolutional siamese fusion networks for object tracking". In \textit{Proc. ICIP} 2018, Athens, Greece, October 7-10, 2018, pages 3718–3722.

\bibitem{CompNet}
Tengfei Zhang, Yue Zhang, Xian Sun, Hao Sun, Menglong Yan, Xue Yang, and Kun Fu. ``Comparison network for one-shot conditional object detection", 2019. [Online]  Available: \url{https://arxiv.org/abs/1904.02317}.

\bibitem{paper24}
Hang Su, Xiatian Zhu, Shaogang Gong. ``Open Logo Detection Challenge". In \textit{Proc. BMVC}, 2018.

\bibitem{52}
Y. Kalantidis, LG. Pueyo, M. Trevisiol, R. van Zwol, Y. Avrithis. ``Scalable Triangulation-based Logo Recognition". In \textit{Proc. ICMR}, Trento, Italy, April 2011.

\bibitem{53}
S. Romberg, L. G. Pueyo, R. Lienhart, and R. Van Zwol. ``Scalable logo recognition in real-world images". in \textit{Proc. ICMR}, 2011.

\bibitem{54}
Simone Bianco, Marco Buzzelli, Davide Mazzini, and Raimondo Schettini. ``Deep learning for logo recognition". \textit{Neurocomputing}, 245:23–30, 2017.

\bibitem{55}
Alexis Joly and Olivier Buisson. ``Logo retrieval with a contrario visual query expansion". in \textit{Proc. ACM Multimedia}, Beijing, China, 2009, pp. 581–584.

\bibitem{56}
Hang Su, Shaogang Gong, and Xiatian Zhu. ``Weblogo-2m: Scalable logo detection by deep learning from the web". In \textit{Proc. ICCV Workshop}, 2017.

\bibitem{57}
Andras Tüzkö, Christian Herrmann, Daniel Manger, and Jürgen Beyerer. ``Open set logo detection and retrieval", 2017. [Online] Available: \url{https://arxiv.org/abs/1710.10891}.

\bibitem{58}
Yuan Liao, Xiaoqing Lu, Chengcui Zhang, Yongtao Wang, and Zhi Tang. ``Mutual enhancement for detection of multiple logos in sports videos". In \textit{Proc. ICCV}, 2017.

\bibitem{44.1}
J. Deng, W. Dong, R. Socher, L.-J. Li, K. Li, and L. Fei-Fei. ``ImageNet: A Large-Scale Hierarchical Image Database". In \textit{Proc. CVPR}, 2009.

\bibitem{47}
Vinyals, Oriol, Blundell, Charles, Lillicrap, Timothy, Kavukcuoglu, Koray, and Wierstra, Daan. ``Matching networks for one shot learning", 2016. [Online] Available: \url{https://arxiv.org/abs/1606.04080}

\bibitem{VOC}
M. Everingham, L. Van Gool, C. K. Williams, J. Winn, and A. Zisserman. ``The Pascal Visual Object Classes (VOC) Challenge". \textit{IJCV}, pages 303–338, 2010.

\bibitem{COCO}
T.-Y. Lin, M. Maire, S. Belongie, J. Hays, P. Perona, D. Ramanan, P. Dollar, and C. L. Zitnick. ``Microsoft COCO: Common objects in context". In \textit{Proc. ECCV}. 2014.

\bibitem{XiangLi}
Xiang Li, Lin Zhang, Yau Pun Chen, Yu-Wing Tai, Chi-Keung Tang. ``One-Shot Object Detection without Fine-Tuning", 2020. [Online] Available: \url{https://arxiv.org/abs/2005.03819}

\end{thebibliography}
\end{document}